\documentclass[11pt,a4paper]{article}
\usepackage[hyperref]{emnlp2020}
\usepackage{times}
\usepackage{latexsym}

\usepackage{graphicx}
\usepackage{url}
\usepackage{hyperref}
\usepackage{algpseudocode}
\usepackage{algorithm}
\usepackage{amsmath}
\usepackage{booktabs}
\usepackage{setspace}
\usepackage{mathtools}
\usepackage{enumitem}

\usepackage{soul}  %
\setul{}{2pt}  %

\usepackage{enumitem}

\usepackage{microtype}

\aclfinalcopy %
\def\aclpaperid{639} %

\def\Snospace~{\S{}}

\usepackage{tikz} %
\usepackage{pgfplots}
   \usepgfplotslibrary{groupplots}

\pgfplotsset{compat=newest}
\usepgfplotslibrary{units}

\usepackage{textcomp}

\usepackage[textsize=small]{todonotes}

\DeclareMathOperator{\emb}{emb}
\DeclareMathOperator{\simX}{sim}  %

\usepackage{xcolor}
\definecolor{myblue}{rgb}{0.07, 0.04, 0.56}
\hypersetup{colorlinks=true,linkcolor=myblue,citecolor=myblue,urlcolor=myblue}

\let\originalleft\left
\let\originalright\right
\renewcommand{\left}{\mathopen{}\mathclose\bgroup\originalleft}
\renewcommand{\right}{\aftergroup\egroup\originalright}

\title{Exploiting Sentence Order in Document Alignment}

\author{Brian Thompson \\
  Johns Hopkins University \\
  {\tt brian.thompson@jhu.edu} \\\And
  Philipp Koehn \\
  Johns Hopkins University \\
  {\tt phi@jhu.edu} \\}

\date{}

\begin{document}
\maketitle
\begin{abstract}

We present a simple
document alignment method that
incorporates sentence order information in both candidate generation and candidate re-scoring.
Our method results in 61\% relative reduction in error 
compared to the best previously published result on the WMT16 document alignment shared task. 
Our method improves downstream MT performance on web-scraped Sinhala--English documents from ParaCrawl,
outperforming the document alignment method used in the most recent ParaCrawl release.
It also outperforms
a comparable corpora method which uses the same multilingual embeddings, demonstrating
that exploiting sentence order is beneficial
even if the end goal is sentence-level bitext.
\end{abstract}

\section{Introduction}

Document alignment is the task of finding parallel document pairs 
(i.e., documents that are translations of each other) in a large collection of documents,
often crawled from the web. 
Aligned documents have historically been used to produce sentence-level machine translation (MT) data,
but there is growing evidence that MT systems should be trained and evaluated using document-level context
\cite{gong-etal-2011-cache, laubli-etal-2018-machine, voita-etal-2019-good, junczys-dowmunt-2019-microsoft}.

We exploit the simple idea that two parallel documents 
should each contain approximately the same information,
\emph{in approximately the same order}.
This idea can be traced back at least to the late 1990s,
when STRAND \cite{Resnik1998ParallelSA} measured how well linearized HTML tags from two documents
could be aligned in order to judge whether two web pages were likely parallel.
However,
more recent work has primarily used unordered representations for documents,
including bags of words or n-gram features and averages of sentence embeddings.

Our method consists of two main parts:
First, we propose a simple method for candidate generation
which embeds documents into a joint semantic embedding space \cite{berry1995using, wmt16st_uedin2},
in a way that preserves some order information in each document. 
This enables candidate generation via fast approximate nearest neighbor search.
Second, we propose re-scoring those candidate pairs by performing sentence alignment and then scoring that alignment
based on (1) the semantic similarity of the resulting aligned sentence pairs;
(2) whether the sentence pairs are in the correct languages; and
(3) the number of inserted/deleted sentences.
Our re-scoring approach seeks to
filter out documents pairs that contain similar information,
but where the order of that information is not consistent between the two documents.

Our method results in a 61\% relative reduction in the false positive rate 
on the WMT16 document alignment shared task
versus the best previously reported method.
Applied to web-scraped Sinhala--English data from ParaCrawl \cite{paracrawl}, 
it improves MT performance by 1.2 BLEU 
over the document alignment method used in the latest ParaCrawl release \cite{wmt16st_uedin1}, 
when both are used with the Vecalign sentence alignment toolkit \cite{vecalign}.

\section{Method}

We follow a 2-stage approach to
consider
the $D_S \times D_T$ possible alignments between $D_S$ source documents and $D_T$ target 
documents:\footnote{We define the source/target such that $D_S > D_T$.}
\begin{enumerate}[noitemsep,topsep=0pt]
\item \textbf{Candidate Generation}: We first find a fixed number, $K$, of target documents 
as potential matches for each source document.
    \item \textbf{Candidate Re-scoring}: We re-score the $D_S \times K$ document pairs 
from part 1 using a more accurate but slower scoring method.  
\end{enumerate}

Both our candidate generation method and
candidate re-scoring method 
explicitly account for the content of a document
as well as the order of that content within the document. 

\subsection{Candidate Generation}\label{vectormethod}

We propose concatenating several sub-vectors---each emphasizing 
a different section of the document---to form a multilingual document vector.
Each sub-vector is the sum of the sentence embeddings for the entire document,
after embeddings are weighted to emphasize a given region of the document and
to de-emphasize boilerplate text (e.g., from navigational buttons, pull-down menus, or headers).

Let $S_n$ for $n \in \{0,...,N{-}1\}$ be the $N$ sentences in a given document.
We compute sub-vectors $V_j$ 
to emphasize uniformly spaced positions $j \in \{0,...,J{-}1\}$ in the document:
\begin{equation}
    V_j = \sum_{n=0}^{N{-}1} \emb(S_n) \; H_j\left( n \right) \; B(S_n) 
\end{equation}
where
$\emb(S_n)$ is the multilingual embedding of sentence $S_n$ (see \autoref{embeddings}),
$H_j(n)$ is a windowing function to emphasise the $j$\textsuperscript{th} region the document (see \autoref{window}),
and
$B(S_n)$ down-weights boilerplate sentences (see \autoref{boilerplate}).

The final document vector $V$ is a concatenation of normalized position-weighted sub-vectors $V_j$.
Candidate document pairs are found by searching for pairs using cosine distance and
approximate nearest neighbor search.
We compare all documents from a given 
webdomain.\footnote{A webdomain is a top-level website (e.g., \href{www.acted.org}{acted.org}).}

\subsubsection{Sentence Embeddings}\label{embeddings}
Function $\emb(S_n)$ maps sentence $S_n$ into a multilingual vector space.
In this work we use LASER embeddings \cite{laser}, 
as the authors provide a pretrained model that works in 93 languages.\footnote{
\href{https://github.com/facebookresearch/LASER}{github.com/facebookresearch/LASER}}
LASER embeddings require a significant amount of storage space,
so for all experiments in this work 
so we project them from their native size of 1024 down to 128 dimensions 
using Principal Component Analysis (PCA), 
as we find this results in a good performance/space trade-off (see \autoref{pcaAppendix}).

\subsubsection{Windowing Function}\label{window}

$H_j(n)$ is a windowing function to emphasize the $j$\textsuperscript{th} region of a document.
If we were to use a simple rectangular window,
then our method would be equivalent to splitting the document into sections
and computing the average sentence embedding for each section.
However, we instead use many smoothed overlapping windows 
in an effort to encode more fine-grained position information into the final vector document vector,
while also 
making the document alignment process more robust to offsets between parallel sentences,
such as in a document pair with
a boilerplate header or advertisement present in one document but not the other.

For our windowing function $H_j(n)$ we select
a modified PERT distribution \cite{vose2000risk} 
with support over $[0, J]$ and mode $\left(\frac{j{+}0.5}{J}\right)N$.
Modified PERT is based on the PERT \cite{malcolm1959application, clark1962letter} distribution, 
but adds a parameter $\gamma$ to control peakedness of the distribution.
PERT is a re-parameterization of the Beta distribution that is defined by the minimum, most likely
and maximum values a variable can take.

We select $J{=}16$ and $\gamma{=}20$ to produce windows that look reasonable to the authors
(see \autoref{pertWindowsAppendix}).
We do not sweep $J$ or $\gamma$, as
we are concerned about overfitting given our small development set (see \autoref{wmt16stats}).

\subsubsection{Boilerplate Down-weighting}\label{boilerplate}

Many `sentences' in web-crawled data are not true sentences,
but boilerplate text such as text of navigational buttons, headers, or pull-down menus.
We explore three methods for down-weighting such boilerplate text: 
\begin{enumerate}[noitemsep,topsep=0pt]
    \item Scaling by the inverse of the log of number of the documents containing a given sentence,
    inspired by IDF \cite{SparckJones:1988:SIT:106765.106782, wmt16st_uedin1}
    \item A more aggressive variant of IDF which scales sentences by the inverse of the (linear, as opposed to log)
    number of documents containing a given sentence, which we denote `LIDF'
    \item Scaling each sentence by its length, in characters, as boilerplate lines are often very short \cite{boilerplate}.
\end{enumerate}

We find that all three boilerplate methods improve candidate generation performance,
but select LIDF 
as it resulted in the best recall performance on our development set
in preliminary experiments.

\subsection{Candidate Re-scoring}\label{scoringmethod}
To re-score a document pair proposed by candidate generation, we perform sentence alignment
and score the quality of the resulting sentence alignment
in order to judge whether the proposed document pair appears to be a good translation pair.
Our goal is to filter out documents pairs that may contain similar information,
but where the order of that information is not consistent
between the two documents, indicating they are not parallel. 

Our proposed document pair scoring function is:
\begin{multline}\label{scoreequation}
S(E,F) = \\
\frac{1}{|a(E,F)|} 
\smashoperator[r]{\sum_{e,f \in a(E,F) }} \simX(e,f) p(L_E|e) p(L_F|f)
\end{multline}
where $a(E,F)$ is the sentence alignment (see \autoref{sec:sentAlignment}) of documents $E$ and $F$;
$\simX(e,f)$ is the cosine similarity between sentences $e$ and $f$;
and $p(L_e|e)$, $p(L_f|f)$ are the probabilities
that sentences $e$, $f$ are in the correct languages $L_E$, $L_F$ (see \autoref{lid}).
To penalize unaligned sentences,
$a(E,F)$ includes insertions/deletions but we define $\simX(e,f)$ to be zero in such cases.

\subsubsection{Sentence Alignment}\label{sec:sentAlignment}

To perform sentence alignment, we use Vecalign \cite{vecalign}.\footnote{
\href{https://github.com/thompsonb/vecalign}{github.com/thompsonb/vecalign}}
Vecalign uses multilingual sentence embeddings to judge sentence similarity, 
in conjunction with a dynamic programming approximation based on fast dynamic time warping \cite{fastdtw}
to approximate a search over the full space of possible sentence alignments in linear time complexity with respect to document length. 
We follow \citet{vecalign} and again use LASER embeddings,
except we project all embeddings down to size 128.

\subsubsection{Language ID}\label{lid}

One artifact of using multilingual sentence embeddings is that they give perfect alignment scores to exact, 
un-translated sentence copies. 
Since automatic language identification (LID) of web data is often erroneous
and not well defined,\footnote{We observe numerous mixed-language documents 
(e.g., main body in one language and the boilerplate in another).}
this can result in un-translated, (near) duplicate documents being found as document pairs. 
We propose to use all sentences (regardless of language) in sentence alignment, 
as we hypothesize that copies provide a strong signal for sentence alignment.
However, when scoring the alignment
we introduce sentence-level LID probabilities
to penalize sentence pairs that are not in the correct languages.

\section{Experiments and Results}\label{eandr}

We evaluate our document alignment method in both high- and low-resource settings.
Note that our method is not trained on any parallel documents,
and is designed to be as language agnostic as possible.
However, it relies on LASER embeddings, which are trained on bitext.
Thus we expect performance to be at least partially a function of the quantity  
of data that LASER is trained on.%
\footnote{For the two languages considered here, 
LASER was trained on much more French--English data 
(8.8M) than Sinhala--English data (796k) \cite{laser}.
This comparison is likely complicated by data quality
(which we generally expect to be higher in higher-resources languages)
and benefits of training in related languages.}
For high-resource, we use the publicly available French--English data released 
for the WMT 2016 shared task on document alignment \cite{buck2016findings}
and evaluate document recall following the shared task.
The shared task provides a strong set of baselines, as 13 different teams contributed at least one submission.
For low-resource, we experiment with Sinhala--English documents extracted from ParaCrawl.
In this setting we do not have gold document alignments,
so we instead evaluate the quality of MT systems trained on the data extracted via document alignment.

We develop and set all parameters using the training data from WMT16 (`WMT16-train')
and then test on the WMT16 test data (`WMT16-test') and the Sinhala--English ParaCrawl data.
Basic statistics for each dataset are shown in \autoref{wmt16stats}.

\begin{table}
\begin{center}
\begin{tabular}{l  r r r } 
\toprule
  &\multicolumn{2}{c}{WMT16}    & ParaCrawl\\ 
  & train & test &  test \\ 
  \midrule
English Docs. & 349k & 682k & 9.68M \\
French Docs.  & 225k & 522k &     - \\
Sinhala Docs. &    - &    - & 1.49M \\
Webdomains    &   49 &  203 &  1721 \\
Gold Pairs    & 1624 & 2402 &     0 \\
 \bottomrule
\end{tabular}
\end{center}
\caption{Counts for WMT16 and ParaCrawl data used in this work.}\label{wmt16stats}
\end{table}

\subsection{Candidate Generation}

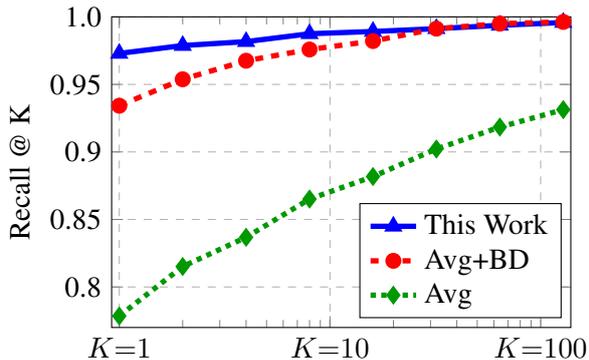
\begin{figure}
\centering
\begin{tikzpicture}
  \begin{axis}[
  log ticks with fixed point,
      grid=major, 
      grid style={dashed,gray!60},
      ylabel=Recall @ K,
      legend style={at={(0.98, 0.03)},anchor=south east},
       width=0.99\linewidth,
       height=5.7cm,
       xtick={1,10,100},
       xticklabels={$K{=}1$, $K{=}10$, $K{=}100$},
       ytick={0.8, 0.85, 0.9, 0.95, 1.0},
       yticklabels={0.8, 0.85, 0.9, 0.95, 1.0},
       legend cell align={left},
       xmode=log,
    xmin=0.92,
    xmax=139,
    ymin=.77,
    ymax=1,
    ]
        \addplot [solid, line width=2.0pt, color=blue, mark=triangle*, mark options={solid}]  table[x=n,y=test_s,col sep=comma] {fr-idf-slots16.csv}; 
    \addlegendentry{This Work}
    \addplot [dashed, line width=2.0pt, color=red, 
    mark=*, mark options={solid}
    ]  
    table[x=n,y=test_s,col sep=comma] {fr-idf-slots1.csv}; 
    \addlegendentry{Avg+BD}
    \addplot [dotted, line width=2.0pt, color=green!60!black, mark=diamond*, mark options={solid}]  table[x=n,y=test_s,col sep=comma] {fr-linear-slots1.csv}; 
    \addlegendentry{Avg}
  \end{axis}
\end{tikzpicture}
\caption{Fraction of the time that a correct document (or near duplicate of it) is found in the top K candidates,
  as a function of K, found by searching document vectors made from average sentence vectors (`Avg'),
  average sentence vectors with boilerplate down-weighting (`Avg+BD'),
  and the proposed method incorporating document order. Results shown on WMT16-test.}\label{recall}
\end{figure}

We find that encoding order in document vectors substantially
reduces the number of candidates, $K$, that must be searched to find the correct document: see \autoref{recall}.
The improvement is largest when a small number of candidates are considered---the proposed method 
approximately halves the number of false positives
between $K{=}1$ and $K{=}10$ compared to the stronger of the two baselines.

\subsection{Document Alignment Recall}\label{wmt16eval}

Within each webdomain, we embed documents as described in \autoref{vectormethod}.
For each French document, we find the top $K{=}32$ candidate translations via approximate nearest neighbor search using FAISS \cite{faiss}. 
We then re-score each candidate pair with \autoref{scoreequation}.
Language ID probabilities are estimated using fastText \cite{joulin2016bag}.\footnote{
\href{https://dl.fbaipublicfiles.com/fasttext/supervised-models/lid.176.bin}
{dl.fbaipublicfiles.com/fasttext/supervised-models/lid.176.bin}
}
We extract the highest scoring document pairs via the greedy search method 
described in \citet{wmt16st_uedin1}.\footnote{\citet{wmt16st_uedin1} 
found that in practice the greedy search outperformed the theoretically optimal Kuhn--Munkres 
algorithm \cite{munkres1957algorithms}.} 

We evaluate document pairs following \citet{buck2016findings}.\footnote{We use their ``soft'' recall,
which gives credit to document pairs for which the English or French document (but not both)
differed from a gold document pair by less than 5\%, as measured by text edit distance.}
The proposed method has a recall of 98.5\%, 
compared to the previous best of 96.2\% (see \autoref{wmt16res}); this corresponds to a 
61\% relative reduction in false positive rate. 
We also try our candidate generation method without rescoring (i.e., $K{=}1$) and find that 
it outperforms prior work, 
but is not as strong as our candidate generation method in conjunction with our candidate re-scoring method. 
For a description of the contrastive methods, see \citet{buck2016findings}.

\begin{table}
\centering
\begin{tabular}{l c c}
\toprule
Method                                  & Recall \\ %
\midrule
\citet{wmt16st_docal}                   & 93.1\% \\ %
\citet{wmt16st_uedin2}                  & 95.0\% \\ %
\citet{wmt16st_novalincs}               & 95.9\% \\ %
\citet{wmt16st_yoda}                    & 96.0\% \\ %
\citet{wmt16st_uedin1}                  & 96.2\% \\ %
\midrule
This Work: Without Re-Scoring           & 97.1\% \\ %
\textbf{This Work: With Re-Scoring}     &  \textbf{ 98.5\%} \\ %
\bottomrule
\end{tabular}\caption{Document recall on WMT16-test, compared to previous best reported results.
The proposed method outperforms prior work, even before re-scoring.
}\label{wmt16res}
\end{table}

\subsection{Impact on Downstream MT}\label{downstream}

We perform document alignment on Sinhala--English documents web-scraped by ParaCrawl.
We apply the same method as in French--English, using the same parameters. 
We compare to document alignment via \citet{wmt16st_uedin1},
followed by sentence alignment using both Vecalign and Hunalign \cite{hunalign}, 
as the latter was used for the most recent ParaCrawl release.

Our document alignment method and Vecalign both use LASER embeddings.
The use of LASER embeddings has been proposed 
for finding 
parallel sentences in comparable corpora (i.e., without doing document alignment),
using a margin-based criterion \cite{artetxe-schwenk-2019-margin}.
Since both methods use the same multilingual embeddings (LASER),
this allows us to determine whether using document-level information
(i.e., performing document alignment and then sentence alignment)
provides better data than simply treating the data as comparable corpora and searching for sentence pairs. 
We refer to this method `LASER-cc.'
For a fair comparison with our document alignment method, we search for sentence pairs within each webdomain.

\begin{table}[t!]
\centering
\begin{tabular}{l r}
\toprule
Method                                    &   \multicolumn{1}{c}{BLEU}  \\ %
\midrule
\textcolor{black}{Buck + Hunalign}        &     8.74 +/- 0.20           \\ %
\textcolor{black}{Buck + Vecalign}        &    10.46 +/- 0.13           \\ %
\textcolor{black}{LASER-cc}               &    10.40 +/- 0.15           \\ %
\midrule
\textcolor{black}{\textbf{This Work + Vecalign}}   &    \textbf{11.62} \textbf{+/-} \textbf{0.09} \\ %
\bottomrule
\end{tabular}\caption{Downstream BLEU (+/- standard deviation for 5 runs)
for the three document alignment + sentence alignment methods compared in this work,
plus the comparable corpora method LASER-cc. `Buck' denotes \citet{wmt16st_uedin1}. 
BLEU shown for best filtering threshold for each method;
see \autoref{bleuplotAppendix} for the results over the entire range of threshold values.
}\label{bleusummary}
\end{table}

For each method of finding parallel sentences, evaluation is the same:
Since the true amount of parallel data is unknown, 
we rank the data from highest to lowest quality following \citet{laserfilt} 
and train systems on a number of different data amounts,
as measured by the number of English words.
We train NMT systems following the WMT19 sentence filtering shared task \cite{wmt19st}. %
Following \citet{vecalign}, we train 5 systems per setting and report both mean and standard deviation BLEU scores.
We report BLEU scores using sacreBLEU \cite{post-2018-call}. %

Results at the best threshold for each method are shown in \autoref{bleusummary}, and results for the full sweep over all thresholds are provided in \autoref{bleuplotAppendix}. 
The proposed method improves downstream MT performance by 1.2 BLEU over \citet{wmt16st_uedin1}, 
when both are used in conjunction with Vecalign,
and 2.9 BLEU over \citet{wmt16st_uedin1} with Hunalign (used in the most recent Paracrawl release).

The proposed method also outperforms the LASER-cc baseline by 1.2 BLEU.
As LASER-cc and the proposed method use the exact same sentence embeddings,
this result shows that incorporating sentence order
not only produces documents that can be used for document-level MT training,
but also results in higher quality sentence pairs.

\section{Related Work}\label{relatedwork}

There is a large amount of prior work in document alignment. 
One of the simplest methods is URL similarity \cite{Resnik1998ParallelSA, chen2000parallel},
although this has been shown to be brittle \cite{tiedemann2011bitext}.
HTML structure \cite{resnik-smith-2003-web, tree} or metadata such as publication date \cite{munteanu-marcu-2005-improving}
is often similar between parallel websites.
However, most more recent work has focused on content similarity %
via bag-of-words or bag-of-ngrams,
using bilingual lexicon \cite{bits, fung-cheung-2004-mining, ion-etal-2011-expectation, wmt16st_uaprompsit, etchegoyhen-azpeitia-2016-portable, azpeitia2019efficient},
machine translation \cite{uszkoreit-etal-2010-large}, or phrase tables \cite{wmt16st_novalincs}. 

Some work has considered high-level order as a filtering step after using a unordered representation to generate candidates: 
\citet{bits} and \citet{wmt16st_ufal} discard n-gram pairs outside a fixed window,
while \citet{uszkoreit-etal-2010-large} filters out documents that have high edit distance between sequences of corresponding n-gram pairs. 
\citet{utiyama-isahara-2003-reliable} and  \citet{Zhang2006AutomaticAO} use sentence similarity and/or number of aligned sentences after performing sentence alignment to score candidate documents.
\citet{guo-etal-2018-effective} score document pairs using the sentence-level nearest neighbor as well as the absolute difference in sentence position between sentence pairs. 
In contrast to these methods, our work considers high-level order in both candidate generation and re-scoring.

\citet{guo-etal-2019-hierarchical} demonstrated neural document embeddings
are effective representations for document alignment.
They trained on millions of document pairs in each specific language pair of interest;
in contrast, this work is much simpler and does not require document-level training data.

\section{Conclusion}

We present a simple but effective method for document alignment.
Our method uses multilingual sentence embeddings 
and explicitly models the \emph{order} of sentences in documents,
in both candidate generation and candidate re-scoring.
Our method outperforms all published results
on the dataset released for the WMT16 shared task
on document alignment.
It also increases downstream MT performance in a low-resource setting
over prior work, including a margin-based comparable corpora method \cite{artetxe-schwenk-2019-margin}.
We use the same embeddings as the comparable corpora method, 
thus the improvement over that method
demonstrates the importance of including sentence order in document alignment,
even when document-level alignments are not required.

\section*{Acknowledgments}
Brian Thompson is supported 
through the National Defense Science and Engineering Graduate (NDSEG) Fellowship Program.

\bibliography{emnlp2020}

\begin{thebibliography}{48}
\expandafter\ifx\csname natexlab\endcsname\relax\def\natexlab#1{#1}\fi

\bibitem[{Artetxe and
  Schwenk(2019{\natexlab{a}})}]{artetxe-schwenk-2019-margin}
Mikel Artetxe and Holger Schwenk. 2019{\natexlab{a}}.
\newblock \href {https://doi.org/10.18653/v1/P19-1309} {Margin-based parallel
  corpus mining with multilingual sentence embeddings}.
\newblock In \emph{Proceedings of the 57th Annual Meeting of the Association
  for Computational Linguistics}, pages 3197--3203, Florence, Italy.
  Association for Computational Linguistics.

\bibitem[{Artetxe and Schwenk(2019{\natexlab{b}})}]{laser}
Mikel Artetxe and Holger Schwenk. 2019{\natexlab{b}}.
\newblock \href {https://doi.org/10.1162/tacl_a_00288} {Massively multilingual
  sentence embeddings for zero-shot cross-lingual transfer and beyond}.
\newblock \emph{Transactions of the Association for Computational Linguistics},
  7:597--610.

\bibitem[{Azpeitia and Etchegoyhen(2016)}]{wmt16st_docal}
Andoni Azpeitia and Thierry Etchegoyhen. 2016.
\newblock \href {https://doi.org/10.18653/v1/W16-2364} {{DOCAL} - vicomtech{'}s
  participation in the {WMT}16 shared task on bilingual document alignment}.
\newblock In \emph{Proceedings of the First Conference on Machine Translation:
  Volume 2, Shared Task Papers}, pages 666--671, Berlin, Germany. Association
  for Computational Linguistics.

\bibitem[{Azpeitia and Etchegoyhen(2019)}]{azpeitia2019efficient}
Andoni Azpeitia and Thierry Etchegoyhen. 2019.
\newblock Efficient document alignment across scenarios.
\newblock \emph{Machine Translation}, pages 1--33.

\bibitem[{Bañón et~al.(2020)Bañón, Chen, Haddow, Heafield, Hoang,
  Esplà-Gomis, Forcada, Kamran, Kirefu, Koehn, Ortiz-Rojas, Pla,
  Ramírez-Sánchez, Sarrías, Strelec, Thompson, Waites, Wiggins, and
  Zaragoza}]{paracrawl}
Marta Bañón, Pinzhen Chen, Barry Haddow, Kenneth Heafield, Hieu Hoang, Miquel
  Esplà-Gomis, Mikel Forcada, Amir Kamran, Faheem Kirefu, Philipp Koehn,
  Sergio Ortiz-Rojas, Leopoldo Pla, Gema Ramírez-Sánchez, Elsa Sarrías,
  Marek Strelec, Brian Thompson, William Waites, Dion Wiggins, and Jaume
  Zaragoza. 2020.
\newblock Paracrawl: Web-scale acquisition of parallel corpora.
\newblock In \emph{Proceedings of the 58th Annual Meeting of the Association
  for Computational Linguistics}. Association for Computational Linguistics.

\bibitem[{Berry and Young(1995)}]{berry1995using}
Michael~W Berry and Paul~G Young. 1995.
\newblock Using latent semantic indexing for multilanguage information
  retrieval.
\newblock \emph{Computers and the Humanities}, 29(6):413--429.

\bibitem[{Buck and Koehn(2016{\natexlab{a}})}]{buck2016findings}
Christian Buck and Philipp Koehn. 2016{\natexlab{a}}.
\newblock \href {https://doi.org/10.18653/v1/W16-2347} {Findings of the {WMT}
  2016 bilingual document alignment shared task}.
\newblock In \emph{Proceedings of the First Conference on Machine Translation:
  Volume 2, Shared Task Papers}, pages 554--563, Berlin, Germany. Association
  for Computational Linguistics.

\bibitem[{Buck and Koehn(2016{\natexlab{b}})}]{wmt16st_uedin1}
Christian Buck and Philipp Koehn. 2016{\natexlab{b}}.
\newblock \href {https://doi.org/10.18653/v1/W16-2365} {Quick and reliable
  document alignment via {TF}/{IDF}-weighted cosine distance}.
\newblock In \emph{Proceedings of the First Conference on Machine Translation:
  Volume 2, Shared Task Papers}, pages 672--678, Berlin, Germany. Association
  for Computational Linguistics.

\bibitem[{Chaudhary et~al.(2019)Chaudhary, Tang, Guzmán, Schwenk, and
  Koehn}]{laserfilt}
Vishrav Chaudhary, Yuqing Tang, Francisco Guzmán, Holger Schwenk, and Philipp
  Koehn. 2019.
\newblock \href {http://www.aclweb.org/anthology/W19-5435} {Low-resource corpus
  filtering using multilingual sentence embeddings}.
\newblock In \emph{Proceedings of the Fourth Conference on Machine Translation
  (Volume 3: Shared Task Papers, Day 2)}, pages 263--268, Florence, Italy.
  Association for Computational Linguistics.

\bibitem[{Chen and Nie(2000)}]{chen2000parallel}
Jiang Chen and Jian-Yun Nie. 2000.
\newblock Parallel web text mining for cross-language ir.
\newblock In \emph{Content-Based Multimedia Information Access-Volume 1}, pages
  62--77. Le Centre de Hautes Etudes Internationales D'Informatique
  Documentaire.

\bibitem[{Clark(1962)}]{clark1962letter}
Charles~E Clark. 1962.
\newblock Letter to the editor -- the pert model for the distribution of an
  activity time.
\newblock \emph{Operations Research}, 10(3):405--406.

\bibitem[{Dara and Lin(2016)}]{wmt16st_yoda}
Aswarth~Abhilash Dara and Yiu-Chang Lin. 2016.
\newblock \href {https://doi.org/10.18653/v1/W16-2366} {{YODA} system for
  {WMT}16 shared task: Bilingual document alignment}.
\newblock In \emph{Proceedings of the First Conference on Machine Translation:
  Volume 2, Shared Task Papers}, pages 679--684, Berlin, Germany. Association
  for Computational Linguistics.

\bibitem[{Espl{\`a}-Gomis et~al.(2016)Espl{\`a}-Gomis, Forcada, Ortiz-Rojas,
  and Ferr{\'a}ndez-Tordera}]{wmt16st_uaprompsit}
Miquel Espl{\`a}-Gomis, Mikel Forcada, Sergio Ortiz-Rojas, and Jorge
  Ferr{\'a}ndez-Tordera. 2016.
\newblock \href {https://doi.org/10.18653/v1/W16-2367} {Bitextor{'}s
  participation in {WMT}{'}16: shared task on document alignment}.
\newblock In \emph{Proceedings of the First Conference on Machine Translation:
  Volume 2, Shared Task Papers}, pages 685--691, Berlin, Germany. Association
  for Computational Linguistics.

\bibitem[{Etchegoyhen and Azpeitia(2016)}]{etchegoyhen-azpeitia-2016-portable}
Thierry Etchegoyhen and Andoni Azpeitia. 2016.
\newblock \href {https://www.aclweb.org/anthology/W16-3412} {A portable method
  for parallel and comparable document alignment}.
\newblock In \emph{Proceedings of the 19th Annual Conference of the {E}uropean
  Association for Machine Translation}, pages 243--255.

\bibitem[{Fung and Cheung(2004)}]{fung-cheung-2004-mining}
Pascale Fung and Percy Cheung. 2004.
\newblock \href {https://www.aclweb.org/anthology/W04-3208} {Mining
  very-non-parallel corpora: Parallel sentence and lexicon extraction via
  bootstrapping and e}.
\newblock In \emph{Proceedings of the 2004 Conference on Empirical Methods in
  Natural Language Processing}, pages 57--63, Barcelona, Spain. Association for
  Computational Linguistics.

\bibitem[{Germann(2016)}]{wmt16st_uedin2}
Ulrich Germann. 2016.
\newblock \href {https://doi.org/10.18653/v1/W16-2368} {Bilingual document
  alignment with latent semantic indexing}.
\newblock In \emph{Proceedings of the First Conference on Machine Translation:
  Volume 2, Shared Task Papers}, pages 692--696, Berlin, Germany. Association
  for Computational Linguistics.

\bibitem[{Gomes and Pereira~Lopes(2016)}]{wmt16st_novalincs}
Lu{\'\i}s Gomes and Gabriel Pereira~Lopes. 2016.
\newblock \href {https://doi.org/10.18653/v1/W16-2369} {First steps towards
  coverage-based document alignment}.
\newblock In \emph{Proceedings of the First Conference on Machine Translation:
  Volume 2, Shared Task Papers}, pages 697--702, Berlin, Germany. Association
  for Computational Linguistics.

\bibitem[{Gong et~al.(2011)Gong, Zhang, and Zhou}]{gong-etal-2011-cache}
Zhengxian Gong, Min Zhang, and Guodong Zhou. 2011.
\newblock \href {https://www.aclweb.org/anthology/D11-1084} {Cache-based
  document-level statistical machine translation}.
\newblock In \emph{Proceedings of the 2011 Conference on Empirical Methods in
  Natural Language Processing}, pages 909--919, Edinburgh, Scotland, UK.
  Association for Computational Linguistics.

\bibitem[{Guo et~al.(2018)Guo, Shen, Yang, Ge, Cer, Hernandez~Abrego, Stevens,
  Constant, Sung, Strope, and Kurzweil}]{guo-etal-2018-effective}
Mandy Guo, Qinlan Shen, Yinfei Yang, Heming Ge, Daniel Cer, Gustavo
  Hernandez~Abrego, Keith Stevens, Noah Constant, Yun-Hsuan Sung, Brian Strope,
  and Ray Kurzweil. 2018.
\newblock \href {https://doi.org/10.18653/v1/W18-6317} {Effective parallel
  corpus mining using bilingual sentence embeddings}.
\newblock In \emph{Proceedings of the Third Conference on Machine Translation:
  Research Papers}, pages 165--176, Brussels, Belgium. Association for
  Computational Linguistics.

\bibitem[{Guo et~al.(2019)Guo, Yang, Stevens, Cer, Ge, Sung, Strope, and
  Kurzweil}]{guo-etal-2019-hierarchical}
Mandy Guo, Yinfei Yang, Keith Stevens, Daniel Cer, Heming Ge, Yun-hsuan Sung,
  Brian Strope, and Ray Kurzweil. 2019.
\newblock \href {https://doi.org/10.18653/v1/W19-5207} {Hierarchical document
  encoder for parallel corpus mining}.
\newblock In \emph{Proceedings of the Fourth Conference on Machine Translation
  (Volume 1: Research Papers)}, pages 64--72, Florence, Italy. Association for
  Computational Linguistics.

\bibitem[{Ion et~al.(2011)Ion, Ceau{\c{s}}u, and
  Irimia}]{ion-etal-2011-expectation}
Radu Ion, Alexandru Ceau{\c{s}}u, and Elena Irimia. 2011.
\newblock \href {https://www.aclweb.org/anthology/W11-1217} {An expectation
  maximization algorithm for textual unit alignment}.
\newblock In \emph{Proceedings of the 4th Workshop on Building and Using
  Comparable Corpora: Comparable Corpora and the Web}, pages 128--135,
  Portland, Oregon. Association for Computational Linguistics.

\bibitem[{Johnson et~al.(2017)Johnson, Douze, and J{\'e}gou}]{faiss}
Jeff Johnson, Matthijs Douze, and Herv{\'e} J{\'e}gou. 2017.
\newblock Billion-scale similarity search with gpus.
\newblock \emph{arXiv preprint arXiv:1702.08734}.

\bibitem[{Joulin et~al.(2016)Joulin, Grave, Bojanowski, and
  Mikolov}]{joulin2016bag}
Armand Joulin, Edouard Grave, Piotr Bojanowski, and Tomas Mikolov. 2016.
\newblock Bag of tricks for efficient text classification.
\newblock \emph{arXiv preprint arXiv:1607.01759}.

\bibitem[{Junczys-Dowmunt(2019)}]{junczys-dowmunt-2019-microsoft}
Marcin Junczys-Dowmunt. 2019.
\newblock \href {https://doi.org/10.18653/v1/W19-5321} {{M}icrosoft translator
  at {WMT} 2019: Towards large-scale document-level neural machine
  translation}.
\newblock In \emph{Proceedings of the Fourth Conference on Machine Translation
  (Volume 2: Shared Task Papers, Day 1)}, pages 225--233, Florence, Italy.
  Association for Computational Linguistics.

\bibitem[{Koehn et~al.(2019)Koehn, Guzmán, Chaudhary, and Pino}]{wmt19st}
Philipp Koehn, Francisco Guzmán, Vishrav Chaudhary, and Juan Pino. 2019.
\newblock \href {http://www.aclweb.org/anthology/W19-5404} {Findings of the
  {WMT} 2019 shared task on parallel corpus filtering for low-resource
  conditions}.
\newblock In \emph{Proceedings of the Fourth Conference on Machine Translation
  (Volume 3: Shared Task Papers, Day 2)}, pages 56--74, Florence, Italy.
  Association for Computational Linguistics.

\bibitem[{Kohlsch\"{u}tter et~al.(2010)Kohlsch\"{u}tter, Fankhauser, and
  Nejdl}]{boilerplate}
Christian Kohlsch\"{u}tter, Peter Fankhauser, and Wolfgang Nejdl. 2010.
\newblock \href {https://doi.org/10.1145/1718487.1718542} {Boilerplate
  detection using shallow text features}.
\newblock In \emph{Proceedings of the Third ACM International Conference on Web
  Search and Data Mining}, WSDM '10, pages 441--450, New York, NY, USA. ACM.

\bibitem[{L{\"a}ubli et~al.(2018)L{\"a}ubli, Sennrich, and
  Volk}]{laubli-etal-2018-machine}
Samuel L{\"a}ubli, Rico Sennrich, and Martin Volk. 2018.
\newblock \href {https://doi.org/10.18653/v1/D18-1512} {Has machine translation
  achieved human parity? a case for document-level evaluation}.
\newblock In \emph{Proceedings of the 2018 Conference on Empirical Methods in
  Natural Language Processing}, pages 4791--4796, Brussels, Belgium.
  Association for Computational Linguistics.

\bibitem[{Le et~al.(2016)Le, Vu, Oberl{\"a}nder, and Bojar}]{wmt16st_ufal}
Thanh~C. Le, Hoa~Trong Vu, Jonathan Oberl{\"a}nder, and Ond{\v{r}}ej Bojar.
  2016.
\newblock \href {https://doi.org/10.18653/v1/W16-2371} {Using term position
  similarity and language modeling for bilingual document alignment}.
\newblock In \emph{Proceedings of the First Conference on Machine Translation:
  Volume 2, Shared Task Papers}, pages 710--716, Berlin, Germany. Association
  for Computational Linguistics.

\bibitem[{Ma and Liberman(1999)}]{bits}
Xiaoyi Ma and Mark~Y. Liberman. 1999.
\newblock Bits: A method for bilingual text search over the web.
\newblock In \emph{In Proceedings of the Machine Translation Summit VII}.

\bibitem[{Malcolm et~al.(1959)Malcolm, Roseboom, Clark, and
  Fazar}]{malcolm1959application}
Donald~G Malcolm, John~H Roseboom, Charles~E Clark, and Willard Fazar. 1959.
\newblock Application of a technique for research and development program
  evaluation.
\newblock \emph{Operations research}, 7(5):646--669.

\bibitem[{Munkres(1957)}]{munkres1957algorithms}
James Munkres. 1957.
\newblock Algorithms for the assignment and transportation problems.
\newblock \emph{Journal of the society for industrial and applied mathematics},
  5(1):32--38.

\bibitem[{Munteanu and Marcu(2005)}]{munteanu-marcu-2005-improving}
Dragos~Stefan Munteanu and Daniel Marcu. 2005.
\newblock \href {https://doi.org/10.1162/089120105775299168} {Improving machine
  translation performance by exploiting non-parallel corpora}.
\newblock \emph{Computational Linguistics}, 31(4):477--504.

\bibitem[{Post(2018)}]{post-2018-call}
Matt Post. 2018.
\newblock \href {https://doi.org/10.18653/v1/W18-6319} {A call for clarity in
  reporting {BLEU} scores}.
\newblock In \emph{Proceedings of the Third Conference on Machine Translation:
  Research Papers}, pages 186--191, Brussels, Belgium. Association for
  Computational Linguistics.

\bibitem[{Resnik(1998)}]{Resnik1998ParallelSA}
Philip Resnik. 1998.
\newblock Parallel strands: A preliminary investigation into mining the web for
  bilingual text.
\newblock In \emph{AMTA}.

\bibitem[{Resnik and Smith(2003)}]{resnik-smith-2003-web}
Philip Resnik and Noah~A. Smith. 2003.
\newblock \href {https://doi.org/10.1162/089120103322711578} {The web as a
  parallel corpus}.
\newblock \emph{Computational Linguistics}, 29(3):349--380.

\bibitem[{Salvador and Chan(2007)}]{fastdtw}
Stan Salvador and Philip Chan. 2007.
\newblock Toward accurate dynamic time warping in linear time and space.
\newblock \emph{Intelligent Data Analysis}, 11(5):561--580.

\bibitem[{Sennrich and Volk(2010)}]{bleualign}
Rico Sennrich and Martin Volk. 2010.
\newblock {MT}-based sentence alignment for {OCR}-generated parallel texts.
\newblock In \emph{The Ninth Conference of the Association for Machine
  Translation in the Americas (AMTA 2010)}.

\bibitem[{Shi et~al.(2006)Shi, Niu, Zhou, and Gao}]{tree}
Lei Shi, Cheng Niu, Ming Zhou, and Jianfeng Gao. 2006.
\newblock \href {https://doi.org/10.3115/1220175.1220237} {A dom tree alignment
  model for mining parallel data from the web.}

\bibitem[{Sparck~Jones(1988)}]{SparckJones:1988:SIT:106765.106782}
Karen Sparck~Jones. 1988.
\newblock \href {http://dl.acm.org/citation.cfm?id=106765.106782} {Document
  retrieval systems}.
\newblock chapter A Statistical Interpretation of Term Specificity and Its
  Application in Retrieval, pages 132--142. Taylor Graham Publishing, London,
  UK, UK.

\bibitem[{Thompson and Koehn(2019)}]{vecalign}
Brian Thompson and Philipp Koehn. 2019.
\newblock \href {https://doi.org/10.18653/v1/D19-1136} {{V}ecalign: Improved
  sentence alignment in linear time and space}.
\newblock In \emph{Proceedings of the 2019 Conference on Empirical Methods in
  Natural Language Processing and the 9th International Joint Conference on
  Natural Language Processing (EMNLP-IJCNLP)}, pages 1342--1348, Hong Kong,
  China. Association for Computational Linguistics.

\bibitem[{Tiedemann(2011)}]{tiedemann2011bitext}
J{\"o}rg Tiedemann. 2011.
\newblock Bitext alignment.
\newblock \emph{Synthesis Lectures on Human Language Technologies},
  4(2):1--165.

\bibitem[{Uszkoreit et~al.(2010)Uszkoreit, Ponte, Popat, and
  Dubiner}]{uszkoreit-etal-2010-large}
Jakob Uszkoreit, Jay Ponte, Ashok Popat, and Moshe Dubiner. 2010.
\newblock \href {https://www.aclweb.org/anthology/C10-1124} {Large scale
  parallel document mining for machine translation}.
\newblock In \emph{Proceedings of the 23rd International Conference on
  Computational Linguistics (Coling 2010)}, pages 1101--1109, Beijing, China.
  Coling 2010 Organizing Committee.

\bibitem[{Utiyama and Isahara(2003)}]{utiyama-isahara-2003-reliable}
Masao Utiyama and Hitoshi Isahara. 2003.
\newblock \href {https://doi.org/10.3115/1075096.1075106} {Reliable measures
  for aligning {J}apanese-{E}nglish news articles and sentences}.
\newblock In \emph{Proceedings of the 41st Annual Meeting of the Association
  for Computational Linguistics}, pages 72--79, Sapporo, Japan. Association for
  Computational Linguistics.

\bibitem[{Varga et~al.(2007)Varga, Hal{\'a}csy, Kornai, Nagy, N{\'e}meth, and
  Tr{\'o}n}]{hunalign}
D{\'a}niel Varga, P{\'e}ter Hal{\'a}csy, Andr{\'a}s Kornai, Viktor Nagy,
  L{\'a}szl{\'o} N{\'e}meth, and Viktor Tr{\'o}n. 2007.
\newblock Parallel corpora for medium density languages.
\newblock \emph{Amsterdam Studies In The Theory And History Of Linguistic
  Science Series 4}, 292:247.

\bibitem[{Voita et~al.(2019)Voita, Sennrich, and Titov}]{voita-etal-2019-good}
Elena Voita, Rico Sennrich, and Ivan Titov. 2019.
\newblock \href {https://doi.org/10.18653/v1/P19-1116} {When a good translation
  is wrong in context: Context-aware machine translation improves on deixis,
  ellipsis, and lexical cohesion}.
\newblock In \emph{Proceedings of the 57th Annual Meeting of the Association
  for Computational Linguistics}, pages 1198--1212, Florence, Italy.
  Association for Computational Linguistics.

\bibitem[{Volk et~al.(2010)Volk, Bubenhofer, Althaus, Bangerter, Furrer, and
  Ruef}]{volk2010challenges}
Martin Volk, Noah Bubenhofer, Adrian Althaus, Maya Bangerter, Lenz Furrer, and
  Beni Ruef. 2010.
\newblock \href
  {http://www.lrec-conf.org/proceedings/lrec2010/pdf/110_Paper.pdf} {Challenges
  in building a multilingual alpine heritage corpus}.
\newblock In \emph{Proceedings of the Seventh conference on International
  Language Resources and Evaluation ({LREC}{'}10)}, Valletta, Malta. European
  Languages Resources Association (ELRA).

\bibitem[{Vose(2000)}]{vose2000risk}
D~Vose. 2000.
\newblock \emph{Risk analysis: a quantitative guide}.
\newblock John Wiley \& Sons.

\bibitem[{Zhang et~al.(2006)Zhang, Wu, Gao, and Vines}]{Zhang2006AutomaticAO}
Ying Zhang, Ke~Wu, Jianfeng Gao, and Phil Vines. 2006.
\newblock Automatic acquisition of {C}hinese-{E}nglish parallel corpus from the
  web.
\newblock In \emph{ECIR}.

\end{thebibliography}
\bibliographystyle{acl_natbib}

\clearpage

\appendix
\onecolumn

\section{Vecalign Speed/Space/Accuracy Trade-off}\label{pcaAppendix}

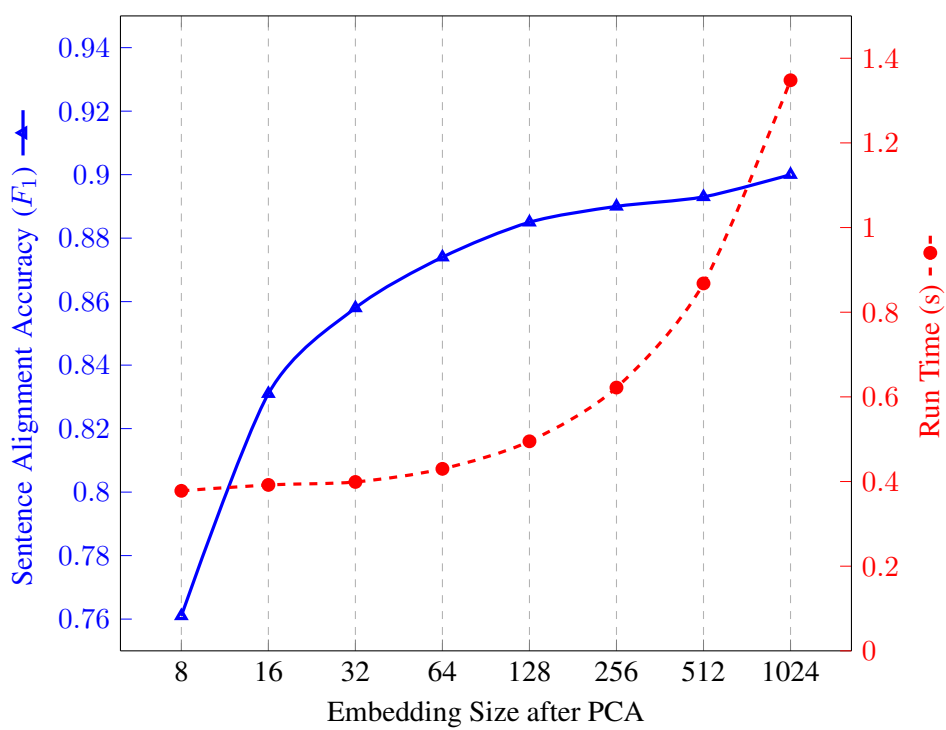
\begin{figure}[h!]
\begin{center}
\begin{tikzpicture}[scale=1.0]
\pgfplotsset{
log ticks with fixed point,
y axis style/.style={
        ylabel style=#1,
        ytick style=#1
    },
}
\begin{axis}[
 xmode=log,
 axis y line*=left,
 y axis style=blue,
 ymin=0.75, ymax=0.95,
 xtick={8, 16, 32, 64, 128, 256, 512, 1024},
 xticklabels={8, 16, 32, 64, 128, 256, 512, 1024},
 xmajorgrids=true,
       grid style={dashed,gray!60},
 xlabel=Embedding Size after PCA,
 ylabel={\hspace{4mm}Sentence Alignment Accuracy ($F_1$) \ref{pgfplots:plot1}},
 ylabel style={text width=7.5cm},
 width=0.7\linewidth,
 height=10cm,
 every y tick label/.append style={blue}  
]
\addplot[smooth,mark=triangle,blue, line width = 1.2,] table[x=size,y=testF1,col sep=comma] {pca.csv};
 \label{pgfplots:plot1}
\end{axis}
\begin{axis}[
 xmode=log,
 axis y line*=right,
 axis x line=none,
 xmajorgrids=true,
       grid style={dashed,gray!60},
 ymin=0.0, ymax=1.5,
 ylabel=Run Time (s) \ref{pgfplots:plot2},
 y axis style=red,
 width=0.7\linewidth,
 height=10cm,
 every y tick label/.append style={red}
]
\addplot[dashed,smooth,mark=*, line width = 1.2,red,mark options={solid}] table[x=size,y=time,col sep=comma] {pca.csv};
\label{pgfplots:plot2}
\end{axis}
\end{tikzpicture}
 \end{center}\caption{$F_1$ (solid blue line) vs time to align (dashed red line) the German--French test set after projecting LASER embeddings to various dimensions using PCA.}\label{fig:pca}
\end{figure}

We experiment with projecting the 1028-dimension LASER embeddings into a lower dimensional space 
using PCA prior to use in Vecalign.
We evaluate sentence alignment accuracy following \citet{vecalign}, 
on the German--French test set released with Bleualign \cite{bleualign},
consisting of manually aligned yearbook articles published in both German and French 
by the Swiss Alpine Club from the Text+Berg corpus \cite{volk2010challenges}.
Accuracy and alignment time for a range of embedding sizes are shown in \autoref{fig:pca}. 
Timing is measured on a laptop with a 1.80GHz i7-8550 CPU. 
We see strong performance ($F_1 > 0.85$)
for embeddings down to size 32,
in conjunction with up to a 70\% reduction in runtime and 97\% reduction in disk space required to store the embeddings.
However, we select a slightly larger dimension of 128 for use in this work.
This projection has minimal impact on sentence alignment accuracy, 
which we expect to have a direct impact on candidate re-scoring performance.
We do not explore the relationship between projected size and candidate generation performance in this work.

\vspace{.5cm}

\clearpage

\section{Modified PERT Window Illustration}\label{pertWindowsAppendix}

\begin{figure}[h!]
\begin{center}
\includegraphics[width=0.6\linewidth]{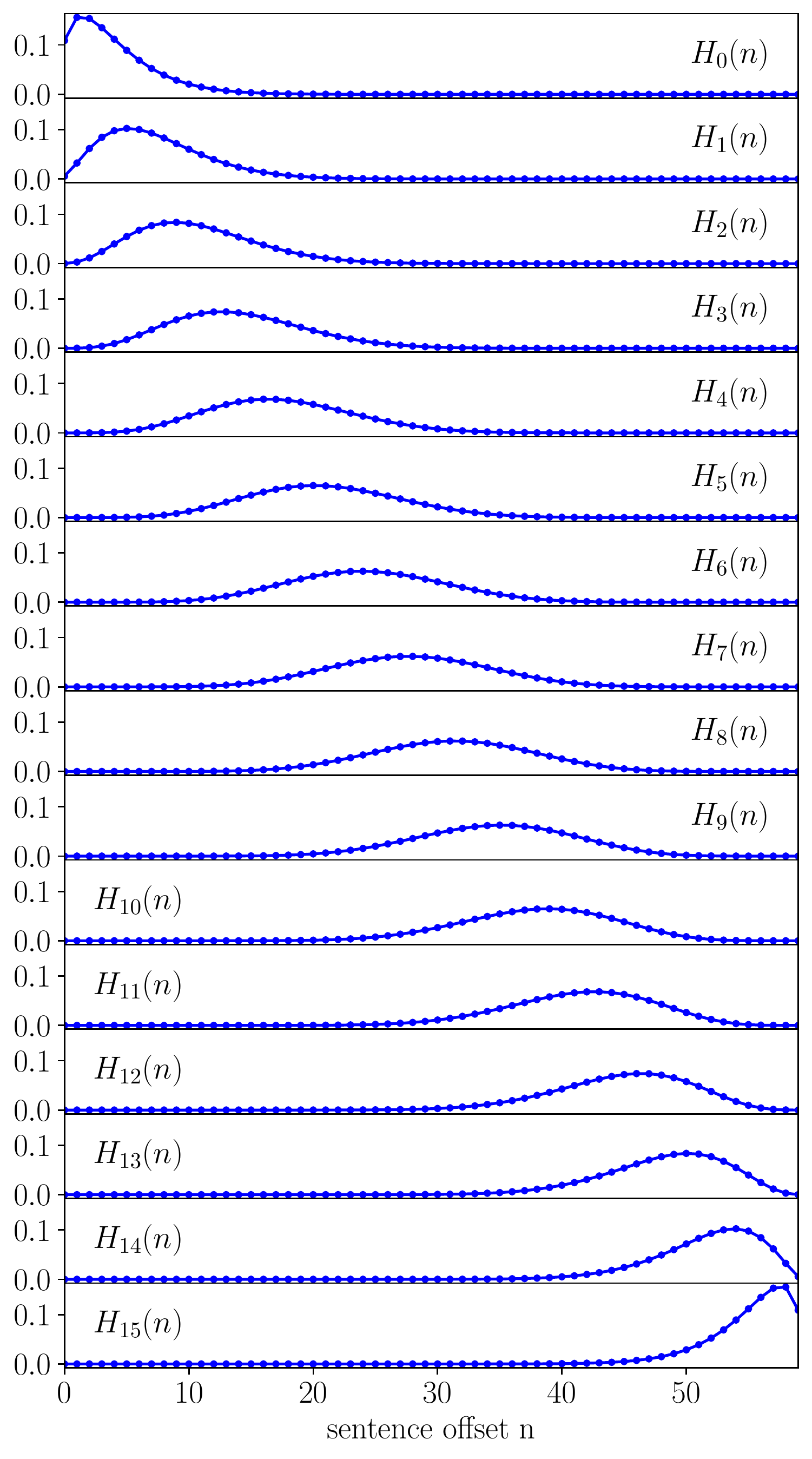} 
\end{center}
\caption{The 16 modified PERT windows used in this work, for an example document containing 60 sentences. 
Each window emphasizes a different region of the document, 
but the regions have substantial overlap in an effort to make the final document vector robust to alignment noise, 
such as offsets caused by a boilerplate header or advertisement present in one document but not the other.
}\label{pertplot}
\end{figure}

\autoref{pertplot} shows the 16 modified PERT windows used in this work, for an example document.
We select $J{=}16$ and $\gamma{=}20$ to produce windows that look reasonable to the authors,
but do not explore sweeping either parameter due to concerns about overfitting on the development set.

\clearpage

\section{Downstream MT Performance for All Thresholds}\label{bleuplotAppendix}

\begin{figure}[h!]
\centering
\begin{tikzpicture}[scale=0.9]
    \begin{axis}[
      ymode=normal,
      ylabel=BLEU,
      xlabel=\# English Words,
      legend cell align={left},
      legend style={at={(0.98, 0.02)},anchor=south east, row sep=3pt},
      grid=major, 
      ymin=2.5,  
      ymax=12,
      xmin=300000,
      xmax=5100000,
      xtick={500000, 1000000, 1500000, 2000000, 2500000, 3000000, 3500000, 4000000,  4500000, 5000000},
      xticklabels={,1M,, 2M,, 3M,, 4M,, 5M},
      ytick={3,4,5,6,7,8,9,10,11,12},
      yticklabels={3,4,5,6,7,8,9,10,11,12},
      scaled x ticks = false,
      grid style={dashed,gray!60},
      width=0.8\linewidth, 
      height=15cm,        
      ]

    \addlegendimage{no markers, white}
    \addlegendentry{\textbf{\ul{Doc. Align + Sent. Align:}}} 

    \addplot[solid, line width=1.5pt, color=blue, mark=., mark options={solid},
        error bars/.cd,
        y dir=both,
        y explicit,
        error bar style={line width=2pt, solid},
        error mark options={
          rotate=90,
          mark size=3pt,
          line width=2pt,
          solid,
        }
    ] table [col sep=comma, x=n, y=mean, y error=std] {bleu_si_vecaligndoc.csv};
    \addlegendentry{\textcolor{blue}{This Work \hspace{0.2mm} + \hspace{0.5mm} Vecalign}};  
      
    \addplot[solid, dash pattern={on 7pt off 2pt on 1pt off 3pt},
    line width=1.5pt, color=red, mark=., mark options={solid},
        error bars/.cd,
        y dir=both,
        y explicit,
        error bar style={line width=2pt, solid},
        error mark options={
          rotate=90,
          mark size=3pt,
          line width=2pt,
          solid,
        }
    ] table [col sep=comma, x=n, y=mean, y error=std] {bleu_si_v3.csv};
    \addlegendentry{\textcolor{red}{\hspace{4.0mm} Buck \hspace{3.2mm} + \hspace{1mm} Vecalign}};  

    \addplot[dashed, line width=1.5pt, color=brown!70!black, mark=., mark options={solid},
        error bars/.cd,
        y dir=both,
        y explicit,
        error bar style={line width=2pt, solid},
        error mark options={
          rotate=90,
          mark size=3pt,
          line width=2pt,
          solid,
        }
    ] table [col sep=comma, x=n, y=mean, y error=std] {bleu_si_v2.csv};
    \addlegendentry{\textcolor{brown!50!black}{\hspace{4.0mm} Buck \hspace{3.2mm} + \hspace{1.0mm} Hunalign}};

    \addlegendimage{no markers, white}
    \addlegendentry{\textbf{\ul{CC Method:}}} 

    \addplot[solid, 
    dash pattern={on 2pt off 2pt},
    line width=1.5pt, color=green!60!black, mark=., mark options={solid},
        error bars/.cd,
        y dir=both,
        y explicit,
        error bar style={line width=2pt, solid},
        error mark options={
          rotate=90,
          mark size=3pt,
          line width=2pt,
          solid,
        }
    ] table [col sep=comma, x=n, y=mean, y error=std] {bleu_si_laser.csv};
    \addlegendentry{\textcolor{green!40!black}{LASER-cc}};  
    
\end{axis}
\end{tikzpicture}
\caption{BLEU scores (mean +/- standard deviation for 5 training runs) for systems trained on
  parallel sentences extracted via several methods,
  over a range of different filtering thresholds.
  `Buck' denotes \citet{wmt16st_uedin1}.
LASER-cc denotes the comparable corpora method of \citet{artetxe-schwenk-2019-margin}.
}\label{bleuplot}
\end{figure}
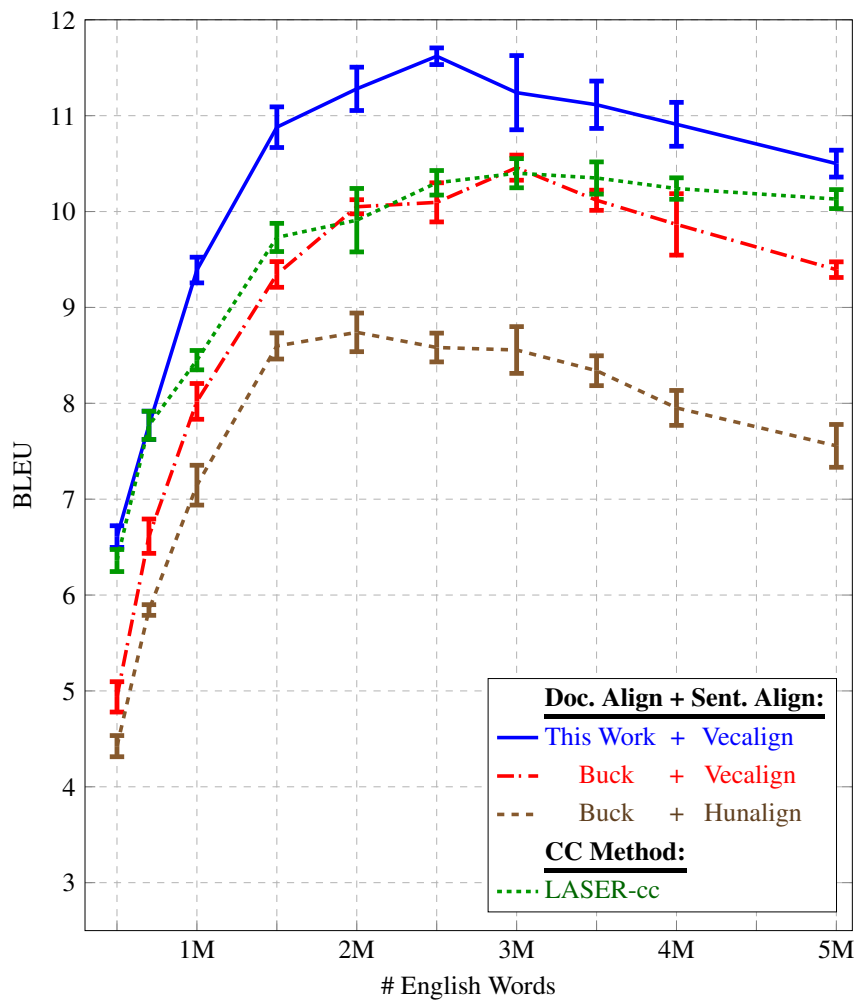

Since the underlying amount of aligned Sinhala--English documents from ParaCrawl is unknown,
in order to evaluate downstream MT performance
we rank the sentence pairs produced by each method
from highest to lowest quality following \cite{laserfilt}
and train each system on many different thresholds.
The thresholds for each method are selected to produce different amounts of data,
which we measure in English words.
Results are shown in \autoref{bleuplot}.

\end{document}